\documentclass{article} 
\PassOptionsToPackage{dvipsnames}{xcolor}
\usepackage{iclr2025_conference,times}

\usepackage{amsmath,amsfonts,bm}

\def\eqref#1{equation~\ref{#1}}

\def\1{\bm{1}}

\DeclareMathAlphabet{\mathsfit}{\encodingdefault}{\sfdefault}{m}{sl}
\SetMathAlphabet{\mathsfit}{bold}{\encodingdefault}{\sfdefault}{bx}{n}

\usepackage{hyperref}
\usepackage{url}
\usepackage{soul}
\usepackage{cleveref}
\usepackage{booktabs}
\usepackage{multicol}
\usepackage{multirow}
\usepackage{makecell}
\usepackage{graphicx}
\usepackage{xspace}

\newcommand{\envsetupgithub}{\url{https://github.com/JetBrains-Research/EnvBench}\xspace}
\newcommand{\envsetuphf}{\url{https://jb.gg/envbench}\xspace}

\newcommand{\benchname}{\textsc{EnvBench}\xspace}

\title{\benchname: A Benchmark for Automated\\Environment Setup}

\author{Aleksandra Eliseeva, 
Alexander Kovrigin, 
Ilia Kholkin\thanks{Work done during internship at JetBrains Research.}, 
Egor Bogomolov, 
Yaroslav Zharov \\
JetBrains Research \\
Correspondence to \href{mailto:alexandra.eliseeva@jetbrains.com}{{\ttfamily alexandra.eliseeva@jetbrains.com}}
}

\iclrfinalcopy 
\begin{document}

\maketitle

\begin{abstract}

Recent advances in Large Language Models (LLMs) have enabled researchers to focus on practical repository-level tasks in software engineering domain. In this work, we consider a cornerstone task for automating work with software repositories---environment setup, \textit{i.e.}, a task of configuring a repository-specific development environment on a system. Existing studies on environment setup introduce innovative agentic strategies, but their evaluation is often based on small datasets that may not capture the full range of configuration challenges encountered in practice. To address this gap, we introduce a comprehensive environment setup benchmark \benchname. It encompasses 329 Python and 665 JVM-based (Java, Kotlin) repositories, with a focus on repositories that present genuine configuration challenges, excluding projects that can be fully configured by simple deterministic scripts. To enable further benchmark extension and usage for model tuning, we implement two automatic metrics: a static analysis check for missing imports in Python and a compilation check for JVM languages. We demonstrate the applicability of our benchmark by evaluating three environment setup approaches, including a simple zero-shot baseline and two agentic workflows, that we test with two powerful LLM backbones, GPT-4o and GPT-4o-mini. The best approach manages to successfully configure 6.69\% repositories for Python and 29.47\% repositories for JVM, suggesting that \benchname remains challenging for current approaches. Our benchmark suite is publicly available at~\envsetupgithub. The dataset and experiment trajectories are available at~\envsetuphf.
\end{abstract}

\section{Introduction}

Recent advances in Large Language Models (LLMs) have enabled their application across many domains, including software engineering~\citep{llm4se-survey}. Their capabilities in reasoning and interaction with external environments~\citep{agents4se-survey-1, agents4se-survey-2}, as well as in efficient processing of large amounts of information~\citep{long-context-survey}, have allowed researchers to tackle practical repository-level software engineering tasks, such as code generation~\citep{repobench, commit0}, code editing~\citep{swebench}, and code understanding~\citep{understand-whole-repo, repoagent, repoqa}.

In this work, we focus on another repository-level task that programmers face regularly---\textit{environment setup}, \textit{i.e.}, configuring the system to work with an arbitrary software project, for instance, a freshly cloned GitHub repository. It usually entails installing the dependencies but might include arbitrary project-specific steps, such as installing additional system packages, setting the correct environment variables, and more.
A well-maintained project should be straightforward to set up, however, in practice, it is not always the case. For instance, setting up the repository is perceived to be the most challenging part of reproducing Natural Language Processing (NLP) research results, according to~\cite{storks2023nlp}, it may take up to several hours.
Similarly, a survey conducted by \citet{aghajani_software_2020} reveals that incomplete documentation of installation, deployment, and release processes is considered a significant issue by 68\% of developers, and 63\% report inappropriate installation instructions as a prevalent concern.
Moreover, automating environment setup could enable scaling of the execution-based benchmarks, which currently often require significant manual effort to select a set of executable repositories~\citep{swebench}.

As of now, few studies have considered environment setup as a standalone task. 
There are numerous works that include environment setup as a part of a larger task---for instance, 
scientific reproduction~\citep{siegel_core-bench_2024, bogin_super_2024} or solving machine learning problems~\citep{tang_ml-bench_2024}.
However, to the best of our knowledge, there are only two works specifically on environment setup concurrent to ours~\citep{beyond-pip-install, you-name-it}. 
While these works represent important progress in automating environment setup, they primarily focus on novel agentic strategies rather than on comprehensive benchmarking. That manifests in a limited number of the software projects and technologies covered in the respective datasets. For instance, the dataset from \cite{beyond-pip-install} features 40 Python repositories, and \cite{you-name-it} includes 10 repositories for each of the considered languages (Python, Java, C, C++, and JavaScript).

Taking this into consideration, we introduce a novel environment setup benchmark---\benchname. It features a diverse set of projects, covering Python (329 repositories) and JVM languages such as Java and Kotlin (665 repositories in total). 
We implement two automatic metrics to verify that the environment is set up correctly --- static analysis to obtain the number of missing imports (\textit{i.e.}, the number of import statements across the codebase that couldn't be resolved via static analysis due to the corresponding package not being installed) for Python and a compilation check for JVM languages. 
We ensure that the projects included in our benchmark present genuine configuration challenges by implementing simple deterministic shell scripts and excluding repositories that can be correctly configured by these scripts alone. Finally, we evaluate three environment setup approaches with two powerful LLMs, GPT-4o and GPT-4o-mini. Our set of baselines includes a simple zero-shot setting, a ReAct~\citep{react} agentic workflow with access to the Bash terminal similar to~\cite{you-name-it} (Bash Agent), and an agentic setting following~\cite{beyond-pip-install} (Installamatic Agent). 

Our findings show that the Bash Agent with GPT-4o achieves the highest success rates, correctly configuring 29.47\% of the JVM repositories and 6.69\% of the Python repositories. Although environment setup for Python remains challenging, LLM-based approaches still reduce the number of missing imports compared to the deterministic script for many repositories, demonstrating their potential. Additionally, we observe that LLM-based approaches that are not explicitly provided with error feedback commonly produce erroneous environment setup scripts. This aligns with previous findings~\citep{beyond-pip-install} that several generation attempts with error feedback significantly improve environment setup capabilities.

Our benchmark suite is publicly available at~\envsetupgithub. The dataset and experiment trajectories are available at~\envsetuphf.

\begin{figure}
    \centering
    \includegraphics[width=\linewidth]{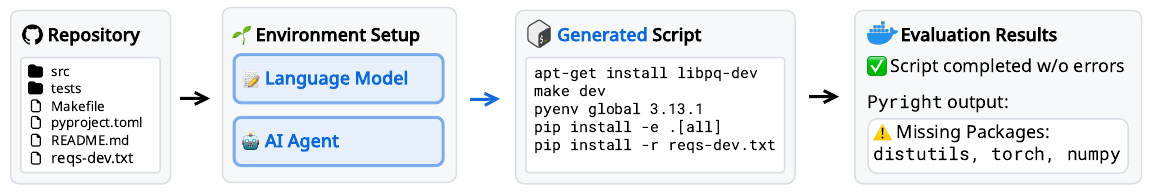}
    \caption{Overview of the workflow with \benchname. The process begins with cloning a target repository.
    Next, the repository is passed as an input to an environment setup approach, which then produces a shell script to set up the repository as an output. Internally, it could be, for instance, a single LLM request or an AI agent building a script dynamically.
    Finally, in our evaluation suite, we execute the produced script and verify the environment is correctly configured through static analysis and compilation checks.}
    \label{fig:envsetup-overview}
\end{figure}

\section{Related Works}

Environment setup, \textit{i.e.}, creating a functioning development environment for a software repository, is a vital task in software development. 
The steps involved vary a lot between programming languages and even between different repositories coming from the same technology stack, thus complicating the automation of this task. Moreover, obtaining a subset of executable repositories from a vast dataset is a common step for execution-based benchmarks. In most recent benchmarks, a check if a repository is executable is done semi-automatically, with a significant amount of manual work required~\citep{swebench,jain2024r2e,tang_ml-bench_2024,commit0}. 

\textbf{Approaches.} 
Given the inherent variability of environment setup of arbitrary repositories, non-ML automatic approaches are limited and usually tied to a specific ecosystem. 
For instance, there are tools that can gather external dependencies from the source code of Python repositories~\citep{gruber_flapy_2023, pipreqs}, yet configuring the system to ensure successful installation is out of their scope. 
Recently, two AI agents for environment setup were introduced. \cite{beyond-pip-install} propose $\mathsf{\textsc{Installamatic}}$ capable of successfully setting up 21 out of 40 considered Python repositories. 
\cite{you-name-it} introduce $\mathsf{\textsc{ExecutionAgent}}$ that correctly configures 33 out of 50 considered repositories across 5 programming languages (Python, Java, C, C++, and JavaScript).

\textbf{Benchmarks.} 
Several recent works have tackled scientific reproduction~\citep{siegel_core-bench_2024, bogin_super_2024} or solving machine learning problems~\citep{tang_ml-bench_2024} --- while both tasks may include environment setup, the evaluation is conducted in an end-to-end fashion, offering little insight into the challenges the current LLMs face during environment setup. To the best of our knowledge, there are two existing benchmarks tailored specifically for the environment setup task --- we'll refer to them by the names of accompanying approaches, $\mathsf{\textsc{Installamatic}}_\text{bench}$~\citep{beyond-pip-install} and $\mathsf{\textsc{ExecutionAgent}}_\text{bench}$~\citep{you-name-it}, respectively. 
$\mathsf{\textsc{Installamatic}}_\text{bench}$ contains 40 Python repositories. \cite{beyond-pip-install} manually inspect the repositories and provide ground truth installation-relevant context and an exemplar Dockerfile for each. The expected output is a Dockerfile, and the success metric is for at least one test to pass with the generated Dockerfile.
$\mathsf{\textsc{ExecutionAgent}}_\text{bench}$ covers 5 programming languages (Python, Java, C, C++, and JavaScript) with 10 repositories for each programming language. \cite{you-name-it} select the repositories with CI logs available, providing the ground truth results of test suite execution for each repository. The expected outputs are both a Dockerfile that specifies system configuration and a shell script that sets up the environment and runs tests. For evaluation, the authors consider three metrics: success build rate, success test rate, and deviation from the ground truth in terms of the number of passing, failing, and skipped tests. The first two metrics require manual inspection. In addition, both works focus on rather popular projects: \cite{beyond-pip-install} consider repositories with at least 1000 stars on GitHub, while \cite{you-name-it} --- with at least 100.

\textbf{Our contributions.}
Compared to existing benchmarks, our benchmark covers a broader range of 994 repositories across three programming languages (Python, Java, Kotlin) and two distinct ecosystems (Python and JVM). We apply more relaxed star filters (minimum 10) and exclude repositories that can be configured using simple deterministic scripts, ensuring genuine environment setup challenges. Unlike benchmarks relying on test execution, we use static analysis (Python) and compilation checks (JVM) to verify successful setup. Similar to~\cite{you-name-it}, we employ shell script output format for the environment setup approaches, however, we provide predefined Dockerfiles to ensure that base system configuration remains consistent across all approaches.

\section{\benchname Benchmark}\label{sec:benchmark}

In this section, we describe \benchname---our benchmark for environment setup task. We consider repositories written in Python or in the JVM-based languages (specifically, Java and Kotlin), representing two popular\footnote{For instance, from GitHub's Octoverse report from October 2024, Python is the most used language on GitHub, and Java is the 4th: \url{https://github.blog/news-insights/octoverse/octoverse-2024/}} yet fundamentally different technology stacks. 
Refer to~\Cref{sec:bench_stats} for additional information about the repositories in our benchmark. 
\benchname is available at~\envsetuphf. Our evaluation suite and other associated code are available at~\envsetupgithub.

\subsection{Task definition}\label{sec:task-definiton}

We present an overview of the expected workflow within our benchmark in~\Cref{fig:envsetup-overview}.

\textbf{Input and output.} In our benchmark, the input for an environment setup approach is the full repository contents, and the expected output is a shell script that configures the repository. 
Both the processing of repository contents and the script generation method are integral to the approach. For example, the process could involve a single LLM request, where the repository context is gathered using a predefined algorithm, or an LLM agent that dynamically explores the repository and executes shell commands via provided tools to generate the script.

\textbf{Evaluation metrics.} 
Given the differences between Python and JVM languages, we implement two distinct metrics to evaluate whether the repository environment was configured correctly. For Python, we run a popular static analysis tool pyright~\citep{pyright} and count the number of reported errors related to missing dependencies (specifically, of {\ttfamily reportMissingImports} type). 
For JVM languages, we try to build the repository via either Gradle ({\ttfamily gradle build} command) or Maven ({\ttfamily mvn compile} command), check if the attempt was successful (for both), and report the number of errors in build tool output (for Maven) . In the process of the benchmark construction, we verify that the included JVM-based repositories use either Gradle or Maven build tools based on the presence of the configuration files.
Both metrics can finish execution with non-zero exit codes if the script for configuring a repository is incorrect.
In most of the considered configurations, our metrics allow both a binary success indicator (zero exit code and zero reported errors) and a continuous measure based on the reported errors per repository that could mitigate the presence of the repositories that are objectively infeasible to set up successfully.

In contrast with our approach, the previous works on environment setup adopted test suite-based metrics~\citep{beyond-pip-install, you-name-it} that can be considered closer to the real use-cases. 
Following~\cite{you-name-it}, execution-based metric for the environment setup could consist of three criteria: successfully building (or installing) the project, being able to run the test suite, and test suite working as expected. Each step depends on all the previous finishing successfully. 
In our formulation, we cover build (installation) step, discovering most problems that environment setup approaches could face except those that only appear during runtime. 
Our metrics are more lightweight compared to test suite-based, enabling us to scale our benchmark while our experiments demonstrate that~\benchname remains challenging even for powerful LLM-based environment setup approaches (refer to~\Cref{sec:results} for the results of our experiments).
As environment setup methods advance, \benchname could be extended with test suite-based metrics, ensuring it remains challenging and closely aligned with real-world use cases.

\subsection{Evaluation suite}\label{sec:eval-suite}

Environment setup inherently means performing actions that modify system configuration. To prevent automated approaches from possibly corrupting a host system, we implement an evaluation suite where solutions for each repository are launched in a Docker container~\citep{docker}. The overall evaluation process for each repository is as follows.
Evaluation suite expects as an input the name of the repository, the revision at which the repository should be considered, and the environment setup shell script for the repository, supposedly, produced by an environment setup approach in advance. 
We clone the repository into a Docker container, execute the provided shell script, and, if the script execution finishes successfully, execute the metric for the corresponding language. 
We release two base Docker images, for Python and for JVM languages, that provide a minimal set of relevant tools. More details about our evaluation suite Docker configuration can be found in~\Cref{sec:docker-environment}.

\subsection{Data collection and filtering}\label{sec:data-collection}

\textbf{Data Collection.} To construct~\benchname, we start by obtaining a diverse list of GitHub repositories using a dedicated tool GitHub Search~\citep{github-search}. Our selection criteria include the primary language of the repository being either Python, Java, or Kotlin, repository having a permissive license, and a set of quality filters designed to exclude projects that might introduce biases~\citep{data-collection-filters}: at least 1000 commits; at least 10 issues; at least 10 contributors; at least 10 stars; last commit made not earlier than January 1st, 2024. That way, we focus on mature and active projects, but include both popular and those that have received less recognition. We then clone the source code of the repositories that are still available at the time of July 2024, which resulted in 2,590 repositories for Python and 1,688 for JVM languages.

\textbf{Filtering by Contents.} We study the configuration files present in the repositories and observe that the majority of Python projects use either pip~\citep{pip} or Poetry~\citep{poetry} to manage the dependencies, and the presence of other dependency managers is negligible. Similarly, the majority of JVM projects use either Gradle~\citep{gradle} or Maven~\citep{maven} build tools. So, as the next step, we filter out the repositories that either do not contain configuration files associated with the respective dependency managers in their root directory or contain several. This way, we avoid including monorepos --- \textit{i.e.}, repositories that contain several different projects from the point of view of configuration --- from our sample, as setting up a monorepo is a more challenging task, where evaluation can be quite ambiguous. After this step, we retain 743 repositories for Python and 1,487 for JVM languages.

Additionally, we filter out repositories that contain configuration files related to Docker~\citep{docker} (\textit{e.g.}, {\ttfamily Dockerfile} or {\ttfamily docker-compose.yml}). We rely on Docker to sandbox evaluation and experiments (more details in~\Cref{sec:eval-suite} and~\Cref{sec:exp-setup}, respectively), and running Docker within a Docker container poses significant technical challenges. Furthermore, Docker is often employed to simplify and encapsulate environment setup, potentially bypassing key challenges. Following this filtering step, we obtain 391 repositories for Python and 977 for JVM languages.

\textbf{Baseline Filtering.} Intuitively, for many repositories environment setup can be as simple as running {\ttfamily pip install -r requirements.txt}. To ensure that our benchmark remains non-trivial, we implement two fully deterministic shell scripts, one for Python and one for JVM languages (refer to~\Cref{sec:baseline-scripts} for details). 
At the core, the algorithms are similar: (1) analyze configuration files to determine the required dependency manager and Python/Java version; (2) verify if the correct Python/Java version is installed, and install it if necessary; and (3) install packages using the specified dependency manager (for Python; for JVM, project builds are excluded as they occur during evaluation).

We run the baseline scripts and assess their performance using the evaluation suite outlined in~\Cref{sec:eval-suite}. For Python, the script successfully (as defined in~\Cref{sec:task-definiton}) sets up 62 repositories (15.9\%), while for JVM languages --- 309 repositories (31.6\%).
From the sample obtained on the previous step, we were unable to process 3 repositories. 
After filtering out these unprocessable repositories, as well as those successfully configured by the baseline scripts, we obtain a final dataset of 329 Python repositories and 665 JVM repositories.

\section{Experimental setup}\label{sec:exp-setup}

\subsection{Dataset \& Metrics}

Our dataset, \benchname, features 329 Python and 665 JVM repositories. Its construction is described in details in~\Cref{sec:benchmark}. 
We implemented two language-specific metrics that rely on either static analysis (for Python) or compilation check (for JVM languages) to confirm if the repository was configured correctly (refer to~\Cref{sec:task-definiton}). These metrics output the number of observed errors, however, depending on a shell script produced by an environment setup approach, they might also exit prematurely with a non-zero exit code. We consider two metrics: \textit{pass@1}, a binary measure of success---where success is defined as both the exit code and reported errors being zero---and \textit{avgErrs}---the average number of reported errors per repository---which quantifies the extent to which the setup was completed. Note that \textit{avgErrs} can only be computed for the repositories where the environment setup script finished execution with a zero exit code.

\subsection{Baselines}

We consider three LLM-based baselines in our experiments. For each baseline, we run experiments with two proprietary LLMs: GPT-4o and GPT-4o-mini. Refer to~\Cref{sec:baselines-implementation} for implementation details of each baseline.

\textbf{Zero-shot LLM.} We construct a prompt with information about a repository (including a directory structure, README contents, and the contents of the configuration files) and system configuration for our evaluation suite (\Cref{sec:eval-suite}) and send a single request to an LLM, asking it to generate a shell script that correctly configures the environment for the given repository.

\textbf{Installamatic Agent.} We consider environment setup agent $\mathsf{\textsc{Installamatic}}$ introduced by \cite{beyond-pip-install}. $\mathsf{\textsc{Installamatic}}$ comprises two stages: (1) a search phase, where the agent explores the repository contents, and (2) a build/repair phase, where the agent generates and tests a Dockerfile, with several regeneration attempts allowed in case of failure. We slightly adapt $\mathsf{\textsc{Installamatic}}$ to match our evaluation setup (refer to~\Cref{sec:eval-suite} for details) by asking the agent to generate a shell script instead of a Dockerfile, providing agent with extra information about the system configuration for our evaluation suite, and crafting a separate prompt for JVM repositories. We also do not allow regeneration attempts and report the results with the first version of the script produced by the agent, which is one of the settings considered by~\cite{beyond-pip-install}. In this formulation, $\mathsf{\textsc{Installamatic}}$ can be considered an extension of Zero-shot LLM that employs an agent for context gathering instead of a predefined prompt template; however, shell script generation is still conducted as a single LLM request.

\textbf{Bash Agent.} We consider a ReAct~\citep{react} agent that approaches the tasks iteratively, producing thoughts and actions at each step based on previous observations. As the available actions, we provide a single {\ttfamily execute\_bash\_command} tool that allows interaction with the system via shell commands. Due to safety considerations, we execute the commands issued by the agent inside a Docker container. Compared to Zero-shot LLM and $\mathsf{\textsc{Installamatic}}$, this baseline unites both context gathering and shell script generation in an iterative and dynamic process fully controlled by an LLM agent.
Additionally, Bash Agent can be considered as a simplified version of the $\mathsf{\textsc{ExecutionAgent}}$ introduced by \cite{you-name-it}, as it is similar, but lacks a few components such as meta-prompting and summarization of the shell commands output.

\section{Results \& Discussion}\label{sec:results}

\begin{table}[h!]
\centering
\begin{tabular}{llcccc}
\toprule
\multirow{2}{*}[-0.75em]{\textbf{Baseline}} & \multirow{2}{*}[-0.75em]{\textbf{Model}} & 
\multicolumn{2}{c}{\textbf{JVM}}
& 
\multicolumn{2}{c}{\textbf{Python}}\\
\cmidrule(lr){3-4}\cmidrule(lr){5-6}
&&\textbf{pass@1 $\mathbf{\uparrow}$}&\makecell{\textbf{avgErrs} $\mathbf{\downarrow}$\\{\scriptsize\textbf{(Maven)}}}&\textbf{pass@1 $\mathbf{\uparrow}$}&\makecell{\textbf{avgErrs} $\mathbf{\downarrow}$\\{\scriptsize\textbf{(overlap)}}}\\
\midrule
\multirow{2}{*}[-0.6em]{Zero-shot LLM} 
& GPT-4o & \makecell{8.57\%\\{\scriptsize 57/665}} & \makecell{480.50\\{\scriptsize 30}} & \makecell{5.47\%\\{\scriptsize 18/329}} & 54.89 \\
& GPT-4o-mini & \makecell{11.13\%\\{\scriptsize 74/665}} & \makecell{202.97\\{\scriptsize 72}} & \makecell{4.56\%\\{\scriptsize 15/329}} & 151.30 \\
\midrule
\multirow{2}{*}[-0.6em]{Installamatic Agent} & GPT-4o & \makecell{1.35\%\\{\scriptsize 9/665}} & \makecell{\textbf{21.43}\\{\scriptsize 65}} & \makecell{4.86\%\\{\scriptsize 16/329}} & 108.93 \\
& GPT-4o-mini & \makecell{3.01\%\\{\scriptsize 20/665}} & \makecell{33.53\\{\scriptsize 32}} & \makecell{2.74\%\\{\scriptsize 9/329}} & 83.57 \\
\midrule
\multirow{2}{*}[-0.6em]{Bash Agent} & GPT-4o & \makecell{\textbf{29.47\%}\\{\scriptsize 196/665}} & \makecell{26.84\\{\scriptsize216}} & \makecell{\textbf{6.69\%}\\{\scriptsize 22/329}} & \textbf{52.00} \\
& GPT-4o-mini & \makecell{26.77\%\\{\scriptsize 178/665}} & \makecell{24.77\\{\scriptsize205}} & \makecell{5.47\%\\{\scriptsize 18/329}} & 79.89 \\

\bottomrule
\end{tabular}
\caption{Main experimental results. \textbf{pass@1} --- percentage of correctly set up repositories, \textit{i.e.}, repositories where our metric returned zero exit code and reported zero issues.
\textbf{avgErrs} --- average number of the reported errors per repository. For JVM, we only report \textbf{avgErrs} for repositories using Maven build tool. Note that \textbf{avgErrs} for each baseline can be calculated only over the repositories where the environment setup script from the corresponding baseline finished with a zero exit code.
For Python, we report \textbf{avgErrs} calculated on 44 repositories where \textit{all} baselines were able to produce scripts finishing with a zero exit code.
For JVM, there are no such repositories, so we specify the number of the repositories across which \textbf{avgErrs} is calcualted for each baseline. The symbol $\uparrow$ indicates that higher values in the current column are better, while $\downarrow$ indicates that lower values are better.}
\label{tab:results}
\end{table}

We present the primary results of our experiments in~\Cref{tab:results}. 
Bash Agent with GPT-4o as a backbone is the best-performing environment setup approach in terms of pass@1 both for JVM and for Python, successfully setting up 29.47\% and 6.69\% of the considered repositories, respectively (with GPT-4o). 
Zero-shot LLM is significantly worse for JVM repositories (11.13\% with GPT-4o-mini) and slightly worse for Python (5.47\% with GPT-4o).
Although Installamatic Agent incorporates more advanced context gathering than Zero-shot LLM, it ranks as the lowest baseline among considered, with 3.01\% pass@1 (with GPT-4o-mini) for JVM and 4.86\% (with GPT-4o) for Python. We hypothesize that the underlying reason might be that the prompts in Installamatic mostly encourage considering natural language documents, and those are likely to be of higher quality in the original dataset considered by~\cite{beyond-pip-install} due to the popularity filter (the average number of stars is 19k as compared to 1.9k in our dataset).

Our second metric, avgErrs, indicates 52.00 missing imports per repository for the best-performing Python baseline, Bash Agent with GPT-4o, and 21.43 errors per repository for Installamatic Agent with GPT-4o for JVM repositories that use Maven build tool. 
However, avgErrs can only be computed in cases when baseline-produced environment setup scripts finished execution without errors.
We study the exit codes of the produced shell scripts for both Python and JVM and observe that the considered baselines can struggle to achieve that, complicating the comparison via avgErrs (see~\Cref{sec:exit_codes} for more details). This problem is more prominent for JVM than for Python.

\begin{table}[ht]
\centering
\begin{tabular}{ll ccc cc}
\toprule
\multirow{2}{*}[-0.6em]{\textbf{Baseline}} 
& \multirow{2}{*}[-0.6em]{\textbf{Model}} 
& \multicolumn{3}{c}{\textbf{Repositories}}
& \multirow{2}{*}{\makecell{\textbf{Avg.}\\\textbf{decrease}} $\mathbf{\uparrow}$}
& \multirow{2}{*}{\makecell{\textbf{Avg.}\\\textbf{increase}} $\mathbf{\downarrow}$}\\
\cmidrule(lr){3-5}
& 
& \textbf{Less} $\mathbf{\uparrow}$ & \textbf{Same} & \textbf{More} $\mathbf{\downarrow}$&&\\
\midrule
\multirow{2}{*}{Zero-shot LLM} 
& GPT-4o  
& \makecell{\textbf{53\%}\\{\scriptsize 76/144}} 
& \makecell{34\%\\{\scriptsize 50/144}}
& \makecell{\textbf{13\%}\\{\scriptsize 18/144}} 
& 59\% & 487\%\\
& GPT-4o-mini  
& \makecell{39\%\\{\scriptsize 67/172}} 
& \makecell{42\%\\{\scriptsize 73/172}}
& \makecell{19\%\\{\scriptsize 32/172}} & 56\% & 589\% \\
\midrule
\multirow{2}{*}{Installamatic Agent} 
& GPT-4o 
& \makecell{42\%\\{\scriptsize 61/145}} & \makecell{34\%\\{\scriptsize 49/145}} & \makecell{24\%\\{\scriptsize 35/145}} 
& \textbf{63\%} & 286\%
\\
& GPT-4o-mini 
& \makecell{31\%\\{\scriptsize 36/115}} & \makecell{46\%\\{\scriptsize 53/115}} & \makecell{23\%\\{\scriptsize 26/115}} & 54\% & 303\%\\
\midrule
\multirow{2}{*}{\makecell{Bash Agent}}    
& GPT-4o
& \makecell{31\%\\{\scriptsize 71/228}} & \makecell{52\%\\{\scriptsize 119/228}}
& \makecell{17\%\\{\scriptsize 38/228}} & 53\% & \textbf{206\%}
\\
& GPT-4o-mini
& \makecell{41\%\\{\scriptsize 92/226}} & \makecell{41\%\\{\scriptsize 93/226}} & \makecell{18\%\\{\scriptsize 41/226}} & 51\%
& 211\%\\
\bottomrule
\end{tabular}
\caption{Comparison of missing imports per repository for the considered baselines on the Python sample, relative to the deterministic script described in~\Cref{sec:data-collection}. \textbf{Less}, \textbf{Same}, and \textbf{More} indicate the number of repositories where the baseline approach resulted in fewer, the same, or more missing imports, respectively. \textbf{Avg. decrease} and \textbf{Avg. increase} columns represent the average percentage reduction or increase in missing imports for repositories where the baseline outperformed (\textbf{Less}) or underperformed the deterministic script (\textbf{More}). Statistics are calculated only on repositories where both the baseline and the deterministic script exited with a zero exit code, with the set of repositories varying for each baseline. The symbol $\uparrow$ indicates that higher values in the current column are better, while $\downarrow$ indicates that lower values are better.}
\label{tab:imports}
\end{table}

To further break down the performance of the environment setup baselines for Python, we compare them with expert-produced scripts as an upper bound and with a simple deterministic script that we implemented during benchmark construction as a lower bound. 
For the former, we consider 30 randomly sampled repositories and observe a significant gap between expert-produced scripts and all the considered baselines (see~\Cref{sec:manual} for more details). Our expert-produced scripts achieve a pass@1 of 66.7\% on that sample as compared to 10.0\% pass@1 achieved by the best baseline. 
For the latter, we report the comparison results in~\Cref{tab:imports}. 
For a relatively small percentage of repositories (from 24\% for Installamatic Agent with GPT-4o to 13\% for Zero-shot LLM with GPT-4o), the scripts produced by the baseline methods result in more missing imports than the deterministic script. However, the average increase in the number of missing imports per repository relative to the deterministic script in this case is significant, at least 200\% for all the considered baselines. These repositories could provide valuable insights into the limitations of current methods and serve as a basis for developing more robust environment setup strategies in future work.

On the other hand, the percentage of the repositories where the scripts produced by the considered baselines outperform the deterministic script is higher for all the considered baselines (from 31\% for Bash Agent with GPT-4o to 53\% for Zero-shot LLM with GPT-4o). In this case, the average decrease in the number of missing imports per repository relative to the deterministic script is around 50\%-60\% for all the considered baselines. 
Overall, while fully correct environment setup for Python remains challenging, the considered baselines show potential compared to a simple deterministic script.

\section{Limitations}
Our work has several important limitations that we list below.

\textbf{Docker Support.} We explicitly exclude projects that require Docker for their setup from our benchmark. While Docker is increasingly common in modern development workflows, including such projects would add complexity to our evaluation setup and metrics. Future work could explore extending our approach to handle a wider range of projects.

\textbf{Data Contamination.} Data contamination is another potential concern, as the LLMs we use were trained on public code repositories that may overlap with our benchmark dataset. This could lead to the models having prior exposure to setup instructions for some of the repositories during pre-training. However, given that environment setup guidelines are often not explicitly documented, and even when they exist, they require advanced reasoning capabilities to interpret and follow correctly, we believe this does not significantly impact our findings. As the construction of our benchmark does not require significant manual effort, it can be updated in the future by incorporating newer repositories, thereby reducing the risk of data contamination as LLM training datasets evolve.

\textbf{OSS Code Quality.} The quality of open-source code and documentation varies significantly across repositories. Some repositories may have incomplete or outdated documentation, making environment setup particularly challenging. Others may be completely invalid or impossible to set up due to missing critical files, broken dependencies, or incompatible configurations. Our results may be influenced by this inherent variability in code quality and repository validity, though this reflects real-world conditions that automated tools need to handle. To mitigate this concern, we employ both a binary success metric and the number of reported errors per repository that quantifies the extent to which the environment setup was completed.
As a part of future work, our benchmark could be further manually verified to identify and remove invalid samples.

\textbf{Static Analysis and Compilation.} Our evaluation relies on static analysis for Python and compilation checks for JVM languages rather than actual execution metrics like test suite runs. While this provides a reasonable proxy for environment setup success, it may miss runtime issues that would only appear during execution. 
However, this approach allows us to evaluate environment setup for a larger number of repositories efficiently while avoiding the complexity of test suite execution, which makes our benchmark more representative and allows possibly reusing the built infrastructure for further research that might involve not only evaluation but training of environment setup approaches. Additionally, we validate the robustness of the proposed metrics by manually implementing setup scripts for 30 randomly sampled Python repositories and studying metrics' behavior (see~\Cref{sec:manual} for more details).

\section{Conclusion}

This work presents \benchname---a benchmark for evaluating automated environment setup methods, addressing the limited scale of previous environment setup datasets by covering 329 Python and 665 JVM repositories. Our evaluation suite is based on static analysis for Python and compilation checks for JVM, enabling a systematic assessment of environment setup strategies.

We evaluate three environment setup methods, including two AI agents, with two powerful LLMs as the backbones. Our results demonstrate that environment setup remains challenging for those methods, with the best-performing method successfully configuring only 29.47\% of JVM and 6.69\% of Python repositories. One key challenge is the generation of erroneous scripts, and we leave further investigation and mitigation to future work.

Our benchmark and the associated code are publicly available, providing a scalable platform for further research. In the future, it could be additionally manually verified and extended to incorporate new software repositories, more programming languages, and runtime-based evaluation metrics. 

\bibliography{main}
\bibliographystyle{iclr2025_conference}

\appendix

\section{Benchmark}

In this section, we provide further details about \benchname, evaluation suite, and the deterministic scripts we used for benchmark construction.

\subsection{Benchmark Statistics}\label{sec:bench_stats}

Additional information about the repositories in our benchmark is presented in~\Cref{tab:bench_stats}.

\begin{table}[ht]
  \centering
  \caption{Statistics for the repositories from our benchmark.}
  \label{tab:bench_stats}
  \begin{tabular}{ccccccc}
    \toprule
    \multirow{2}{*}{\textbf{Language}} & \multicolumn{2}{c}{\textbf{General Statistics}} & \multicolumn{4}{c}{\textbf{Dependency Managers Distribution}}\\
    \cmidrule(lr){2-3} \cmidrule(lr){4-7}
    & \textbf{Avg. Stars} & \textbf{Avg. \# Files} & \textbf{Pip} & \textbf{Poetry} & \textbf{Gradle} & \textbf{Maven} \\
    \midrule
    Python   & 1469 & 779 & \makecell{82.06\%\\{\scriptsize 270/329}} & \makecell{17.94\%\\{\scriptsize 59/329}} & --  & --  \\
    JVM      &  2079 & 2647 & -- & -- & \makecell{59.70\%\\{\scriptsize 397/665}} & \makecell{40.30\%\\{\scriptsize 268/665}} \\
    \bottomrule
  \end{tabular}
\end{table}

\subsection{Docker Environment}\label{sec:docker-environment}

We use Docker~\citep{docker} for both evaluation suite and experiments to safely isolate the execution of LLM-produced scripts from the host system. We implement two Docker environments, one for Python and one for JVM languages, where we preinstall commonly needed tools.
We use preconfigured universal Docker images since it saves time by having standard tools preinstalled, ensures that all approaches operate in the same reproducible environment, and mitigates common issues (\textit{e.g.}, our preliminary experiments showed that all the considered approaches struggle to install Android SDK if it is not available on the system beforehand). 
We base the Docker images on \texttt{ubuntu:22.04} and use non-interactive mode for all tools. The Dockerfiles are available in our GitHub repository: \envsetupgithub.

\textbf{Python Environment:} We preinstall:
\begin{itemize}
    \item pyenv with Python 3.8-3.13
    \item Poetry for dependency management
    \item uv for package installation
    \item pyright for static analysis
    \item pipenv
    \item conda/miniconda
\end{itemize}

\textbf{JVM Environment:} We preinstall:
\begin{itemize}
    \item sdkman with Java 11.0.20-tem
    \item Maven
    \item Gradle
    \item Node.js and npm
    \item Android SDK
\end{itemize}

\pagebreak 
\subsection{Deterministic Scripts}\label{sec:baseline-scripts}

We implement two deterministic scripts that can handle the simplest environment setup cases.
The scripts rely on repository contents to determine the required dependency manager and Python/Java version requirements. The exact scripts are available in our GitHub repository: \envsetupgithub.

\textbf{Python Script:}
\begin{itemize}
    \item Detects environment type by checking for environment.yml (Conda), uv.lock (uv), or poetry.lock (Poetry)
    \item Creates and activates the appropriate virtual environment
    \item Installs dependencies by searching for requirements.txt, setup.py, pyproject.toml, setup.cfg, or Pipfile
    \item Exits with error code if no recognized configuration is found
\end{itemize}

\textbf{JVM Script:}
\begin{itemize}
    \item Detects build system by checking for pom.xml (Maven) or build.gradle (Gradle)
    \item Determines Java version from build files or defaults to Java 11
    \item Runs appropriate build command (mvn install or gradle build)
    \item Handles common build flags like skipping tests or offline mode
\end{itemize}

\section{LLM Baselines Implementation Details}\label{sec:baselines-implementation}

In this section, we share details on the implementations of the considered LLM-based environment setup baselines. The implementations are available in our repository: \envsetupgithub.

For \textbf{Zero-shot LLM}, we first follow predefined steps to collect relevant context for each repository.
Specifically, for both languages, we provide the following information: directory structure (\texttt{tree}, \texttt{ls -R}), documentation contents (README, installation guides, Markdown files), environment information (Dockerfile), deterministic script (described in~\Cref{sec:baseline-scripts}) for corresponding language as an example.
For Python, we provide: common configuration files (\texttt{setup.py}, \texttt{pyproject.toml}), dependency specifications (\texttt{requirements.txt}), Python version requirements (if present in configuration files), contents of \texttt{\_\_init\_\_.py} files.
For JVM, we provide: build files (\texttt{pom.xml}, \texttt{build.gradle}, \texttt{settings.gradle}), dependency and lock files, Java version requirements (if present in build files), build tool wrapper scripts, \texttt{module-info.java} files.

After collecting the context, we employ it to prompt LLM to generate an environment setup shell script.

\textbf{Bash Agent} is a  ReAct~\citep{react} agent that is provided access to a terminal of a Docker container to perform environment setup (refer to~\Cref{sec:docker-environment} for details about Docker configuration). The agent is equipped with a single {\ttfamily execute\_bash\_command} tool that accepts a command and returns stdout contents and stderr contents after the command execution.
We use LangGraph~\citep{langgraph} framework to implement Bash Agent. The agent is allowed up to 30 iterations, and the execution finishes prematurely if the LLM does not use the tool in its response. We set a 360 seconds timeout for the execution of each issued command. We allow up to 5000 characters in the output of each command to avoid overly long non-informative contexts and return the first and the last half of the output in case it exceeds this value. To obtain a resulting shell script, we include all the executed commands that finished with exit code zero.

For \textbf{Installamatic Agent}, we follow the original setting~\citep{beyond-pip-install} excluding the repair stage: the agent is allowed one full \textit{Documentation Gathering} stage iteration and one full \textit{Dockerfile Build} stage iteration. During \textit{Documentation Gathering} stage, Installamatic Agent explores the repository via given tools until {\ttfamily finish\_search} tool is called, and the expected output is the list of the files considered to be installation-relevant. During \textit{Dockerfile Build} stage, the agent is allowed to explore installation-relevant files via the same set of tools until {\ttfamily submit\_summary} tool is called; via this tool, the agent produces a natural language summary of the information required to set up the current repository, and afterwards, the agent generates a Dockerfile. We reuse the original prompts, however, reformulate the task to generate a shell script instead, and extend the prompts with information about our evaluation suite Docker configuration. We use LangGraph~\citep{langgraph} framework to implement Installamatic Agent.
\section{Evaluation Results}

In this section, we provide additional results from our experiments.

\subsection{Case Study}

\begin{figure}[h]
    \centering
    \includegraphics[width=\textwidth]{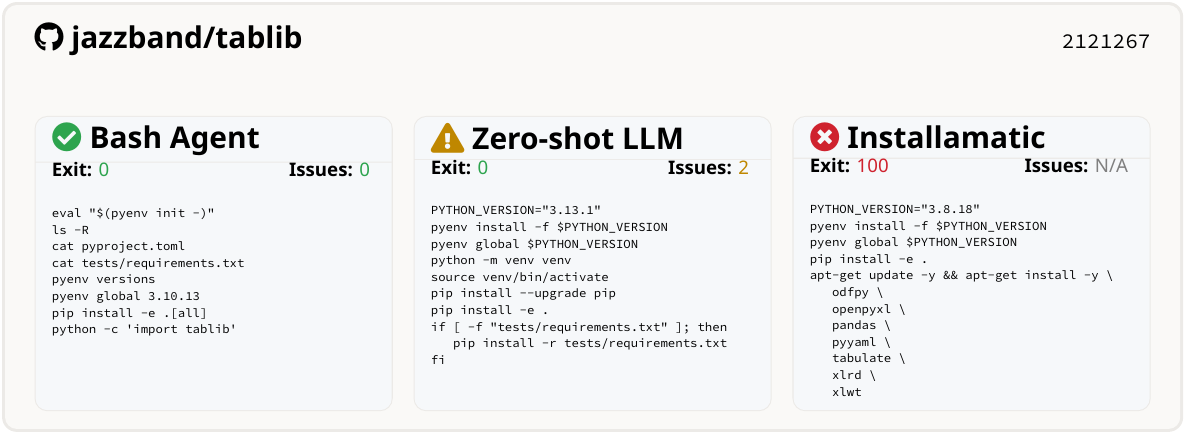}
    \caption{Baselines comparison for the Tablib repository. GPT-4o-mini is used for all baselines. The commented lines have been removed for brevity.}
    \label{fig:python-eval-comparison}
\end{figure}

In this section, we present examples of the scripts generated by the agents and the zero-shot LLM.
In \Cref{fig:python-eval-comparison}, we compare the performance of the baselines for the Python project \href{https://github.com/jazzband/tablib}{\texttt{tablib}} that is a format-agnostic tabular dataset library.

The \texttt{tablib} repository represents a typical Python project setup scenario where:
\begin{itemize}
    \item No explicit installation instructions are provided in the README
    \item No separate \texttt{setup.py} or installation script exists
    \item Dependencies and project metadata are managed through \texttt{pyproject.toml}
    \item Optional dependencies are defined as "extras" that can be installed with \texttt{.[group\_name]}
\end{itemize}

This common configuration requires the installation tool to properly parse the \texttt{pyproject.toml} file to determine Python version requirements and dependencies.

The Bash Agent successfully installs the package by:
\begin{itemize}
    \item Reading \texttt{pyproject.toml} to determine Python version requirements ($\geq$ 3.9)
    \item Installing all optional dependencies via \texttt{.[all]}, which includes all file format support
    \item Using minimal necessary steps without superfluous operations
\end{itemize}

The Zero-shot LLM approach finishes without errors but misses optional dependencies:
\begin{itemize}
    \item Uses Python 3.13.1, which is compatible ($\geq$ 3.9)
    \item Adds unnecessary complexity with virtual environment creation and pip upgrade
    \item Locates and installs test requirements that are not needed in this case
\end{itemize}

The Installamatic Agent fails (exit code 100) by:
\begin{itemize}
    \item Using Python 3.8.18, which violates the package's requirement of $\geq$ 3.9
    \item Installing dependencies through apt instead of using pip, which results in a non-zero exit code
\end{itemize}

This example illustrates common challenges in automated environment setup for Python projects. Even with a relatively standard project structure using modern tooling (\texttt{pyproject.toml}), multiple failure points exist around version compatibility, dependency resolution, and installation method selection. The lack of standardized installation procedures across Python projects, combined with the variety of dependency management approaches, makes automated setup a complex task requiring careful consideration of project-specific requirements and configurations.

\subsection{Costs}
\Cref{tab:costs} presents the token usage and cost statistics for our experiments, calculated using OpenAI API prices as of February 6, 2025. Analysis of these statistics reveals two key patterns.

\textbf{Agent vs Zero-shot Token Usage.} Agent-based approaches consume 5-10x more tokens compared to zero-shot LLM approaches, due to their need to reason about the environment state and process command outputs.
    
\textbf{Language-specific Differences.} JVM projects require significantly more tokens than Python projects across all approaches.
This disparity primarily stems from the more verbose build and dependency resolution logs produced by JVM environments.

Despite these variations in token consumption, the overall costs remain reasonable - even the most token-intensive agent-based approach averages only \$0.25 per repository.

\begin{table}[h]
    \centering
    \begin{tabular}{cllcccc}
    \toprule
    & \multirow{2}{*}{\textbf{Baseline}} & \multirow{2}{*}{\textbf{Model}} & \multicolumn{2}{c}{\textbf{\# tokens}} & \multicolumn{2}{c}{\textbf{Cost}}\\ 
    \cmidrule{4-5}\cmidrule{6-7}
    &  &  & \textbf{Avg.} & \textbf{Total} & \textbf{Avg.} & \textbf{Total}\\ 
    \midrule
    \multirow{6}{*}{\makecell{\rotatebox[origin=c]{90}{\textbf{JVM}}}}
    & \multirow{2}{*}{Zero-shot LLM} & GPT-4o & 15.6k & 10.1M & \$0.042 & \$26.98\\ 
                                     & & GPT-4o-mini & 15.5k & 10.2M & \$0.002 & \$1.60\\
    \cmidrule{2-7}
    & \multirow{2}{*}{\makecell[l]{Bash Agent}} & \makecell[l]{GPT-4o} &\makecell{77k}&\makecell{51M}&\makecell{\$0.20}&\makecell{\$132.8}\\ 
    & & \makecell[l]{GPT-4o-mini} &\makecell{135k}&\makecell{90M}&\makecell{\$0.02}&\makecell{\$13.8}\\
    \cmidrule{2-7}
    & \multirow{2}{*}{Installamatic Agent}
    & \makecell[l]{GPT-4o} &\makecell{98k}&\makecell{65M}&\makecell{\$0.25}&\makecell{\$168.7}\\ 
    & & \makecell[l]{GPT-4o-mini} &\makecell{56k}&\makecell{37M}&\makecell{\$0.01}&\makecell{\$6.1}\\
    \midrule
    \multirow{6}{*}{\rotatebox[origin=c]{90}{\textbf{Python}}}
    & \multirow{2}{*}{Zero-shot LLM} 
    & \makecell[l]{GPT-4o} & 11.0k & 3.6M & \$0.030 & \$10.00\\ 
    & & \makecell[l]{GPT-4o-mini} & 10.8k & 3.6M & \$0.002 & \$0.57\\
    \cmidrule{2-7}
    & \multirow{2}{*}{\makecell{Bash Agent}}  
    & \makecell[l]{GPT-4o} &\makecell{59k}&\makecell{18M}&\makecell{\$0.15}&\makecell{\$47.4}\\ 
    & & \makecell{GPT-4o-mini} &\makecell{97k}&\makecell{32M}&\makecell{\$0.01}&\makecell{\$4.9}\\
    \cmidrule{2-7}
    & \multirow{2}{*}{Installamatic Agent}
    & \makecell[l]{GPT-4o} &\makecell{57k}&\makecell{19M}&\makecell{\$0.15}&\makecell{\$50.1}\\ 
    & & \makecell[l]{GPT-4o-mini} &\makecell{37k}&\makecell{12M}&\makecell{\$0.01}&\makecell{\$2.0}\\
    \bottomrule
    \end{tabular}
    \caption{Usage statistics. Avg. refers to average number per one repository.}
    \label{tab:costs}
\end{table}

\subsection{Exit Codes}~\label{sec:exit_codes}

\begin{table}[ht]
\centering
\begin{tabular}{llcccc}
\toprule
\multirow{2}{*}{\textbf{Baseline}} 
& \multirow{2}{*}{\textbf{Model}}
& \multicolumn{2}{c}{\textbf{Exit Codes (JVM)}}
& \multicolumn{2}{c}{\textbf{Exit Codes (Python)}}\\
\cmidrule(lr){3-4}\cmidrule(lr){5-6}
&
& \makecell{\textbf{Zero} $\mathbf{\uparrow}$}
& \makecell{\textbf{Non-Zero} $\mathbf{\downarrow}$}
& \makecell{\textbf{Zero} $\mathbf{\uparrow}$}
& \makecell{\textbf{Non-Zero} $\mathbf{\downarrow}$}
\\
\midrule
Deterministic Script & --- 
& \makecell{\textbf{98.65\%}\\{\scriptsize 656/665}} 
& \makecell{\textbf{1.35\%}\\{\scriptsize 9/665}} 
& \makecell{71.43\%\\{\scriptsize 235/329}}
& \makecell{28.57\%\\{\scriptsize 94/329}}\\
\midrule
\multirow{2}{*}{Zero-shot LLM} & GPT-4o 
& \makecell{18.80\%\\{\scriptsize 125/665}} 
& \makecell{81.20\%\\{\scriptsize 540/665}} 
& \makecell{47.42\%\\{\scriptsize 156/329}} 
& \makecell{52.58\%\\{\scriptsize 173/329}}\\
& GPT-4o-mini 
& \makecell{27.97\%\\{\scriptsize 186/665}} 
& \makecell{72.03\%\\{\scriptsize 479/665}}
& \makecell{55.62\%\\{\scriptsize 183/329}} 
& \makecell{44.38\%\\{\scriptsize 146/329}}\\
\midrule
\multirow{2}{*}{Installamatic Agent} 
& GPT-4o 
& \makecell{33.83\%\\{\scriptsize 225/665}}
& \makecell{66.17\%\\{\scriptsize 440/665}}
& \makecell{50.76\%\\{\scriptsize 167/329}} 
& \makecell{49.24\%\\{\scriptsize 162/329}}\\
& GPT-4o-mini 
& \makecell{27.82\%\\{\scriptsize 185/665}}
& \makecell{72.18\%\\{\scriptsize 480/665}} 
& \makecell{41.64\%\\{\scriptsize 137/329}} 
& \makecell{58.36\%\\{\scriptsize 192/329}}\\
\midrule
\multirow{2}{*}{\makecell{Bash Agent}}    
& \makecell[l]{GPT-4o} 
& \makecell{82.26\%\\{\scriptsize 547/665}}
& \makecell{17.74\%\\{\scriptsize 118/665}}
& \makecell{\textbf{94.22\%}\\{\scriptsize 310/329}}
& \makecell{\textbf{5.78\%}\\{\scriptsize 19/329}}\\
& \makecell[l]{GPT-4o-mini}
& \makecell{77.29\%\\{\scriptsize 514/665}}
& \makecell{22.71\%\\{\scriptsize 151/665}}
& \makecell{93.31\%\\{\scriptsize 307/329}}
& \makecell{6.69\%\\{\scriptsize 22/329}}\\
\bottomrule
\end{tabular}
\caption{Distribution of the exit codes for the environment setup scripts produced by the considered baselines and the deterministic scripts described in~\Cref{sec:data-collection}.
Zero exit code indicates that the environment setup script finished without errors, however, it is not enough to consider a repository to be set up correctly. The number of successfully set up repositories is reported in~\Cref{tab:results}. By design of our benchmark, the deterministic scripts successfully set up \textit{none} of the repositories.
The symbol $\uparrow$ indicates that higher values in the current column are better, while $\downarrow$ indicates that lower values are better.
}
\label{tab:exit_codes}
\end{table}

We report the results of the considered baselines in terms of exit codes of the produced environment setup scripts in~\Cref{tab:exit_codes}. Here, we also include the results from the simple deterministic scripts described in~\Cref{sec:data-collection} for comparison. We observe that the scripts finishing execution prematurely with a non-zero exit code is a relatively frequent issue for the considered baselines.
For JVM, the deterministic script failed for only 1.35\% of repositories, while LLM-based scripts showed higher failure rates, ranging from 17.74\% (Bash Agent with GPT-4o) to 81.20\% (Zero-shot LLM with GPT-4o). For Python, the deterministic script failed for 28.57\% of repositories, but Bash Agent with GPT-4o reduced this to 5.78\%, outperforming it by 22.79\%.

We qualitatively explore a small sample of the generated environment setup scripts to identify possible root causes. 
For Python, the failures are at times due to the missing system dependencies or a mismatch in Python versions (\textit{e.g.}, if a project relies on an older Python version than the one used on the system and vice versa). Similarly, mismatch in Java versions is a common cause for a non-zero exit code for JVM environment setup scripts. For both languages, the most common reason why the scripts produced by Zero-shot LLM and Installamatic Agent fail is the usage of tools that are unavailable on the system. Compared to those two approaches, Bash Agent can receive immediate error feedback and either install the required tool or consider using another. However, we still observe that it sometimes fails to recover and continues trying to attempt the same failing action repeatedly.

\subsection{Shell Scripts Analysis}

\begin{table}[h!]
\centering
\begin{tabular}{llcccc}
\toprule
\multirow{2}{*}{\textbf{Baseline}} & \multirow{2}{*}{\textbf{Model}} & \multicolumn{2}{c}{\textbf{Avg. \# Lines}} & \multicolumn{2}{c}{\textbf{Avg. Execution Time}} \\
\cmidrule(lr){3-4}\cmidrule(lr){5-6}
& & \textbf{JVM} & \textbf{Python} & \textbf{JVM} & \textbf{Python}\\
\midrule
\multirow[t]{2}{*}{Zero-shot LLM} & GPT-4o & 53.36 & 56.78 & 176.22 & 302.35 \\
& GPT-4o-mini & 33.59 & 40.46 & 221.09 & 294.32 \\
\midrule
\multirow[t]{2}{*}{Installamatic Agent} & GPT-4o & 26.83 & 29.08 & 97.22 & 270.95 \\
 & GPT-4o-mini & 33.71 & 22.13 & 221.08 & 225.94 \\
\midrule

 \multirow[t]{2}{*}{Bash Agent} & GPT-4o & 13.09 & 9.64 & 200.55 & 180.11 \\
 & GPT-4o-mini & 18.30 & 14.64 & 242.84 & 203.35 \\
\bottomrule
\end{tabular}
\caption{Statistics for the environment setup shell scripts produced by the environment setup baselines. Average execution time is reported in seconds.}\label{tab:script-stats}
\end{table}

\begin{figure}[h!]
    \centering
\includegraphics[width=0.8\linewidth]{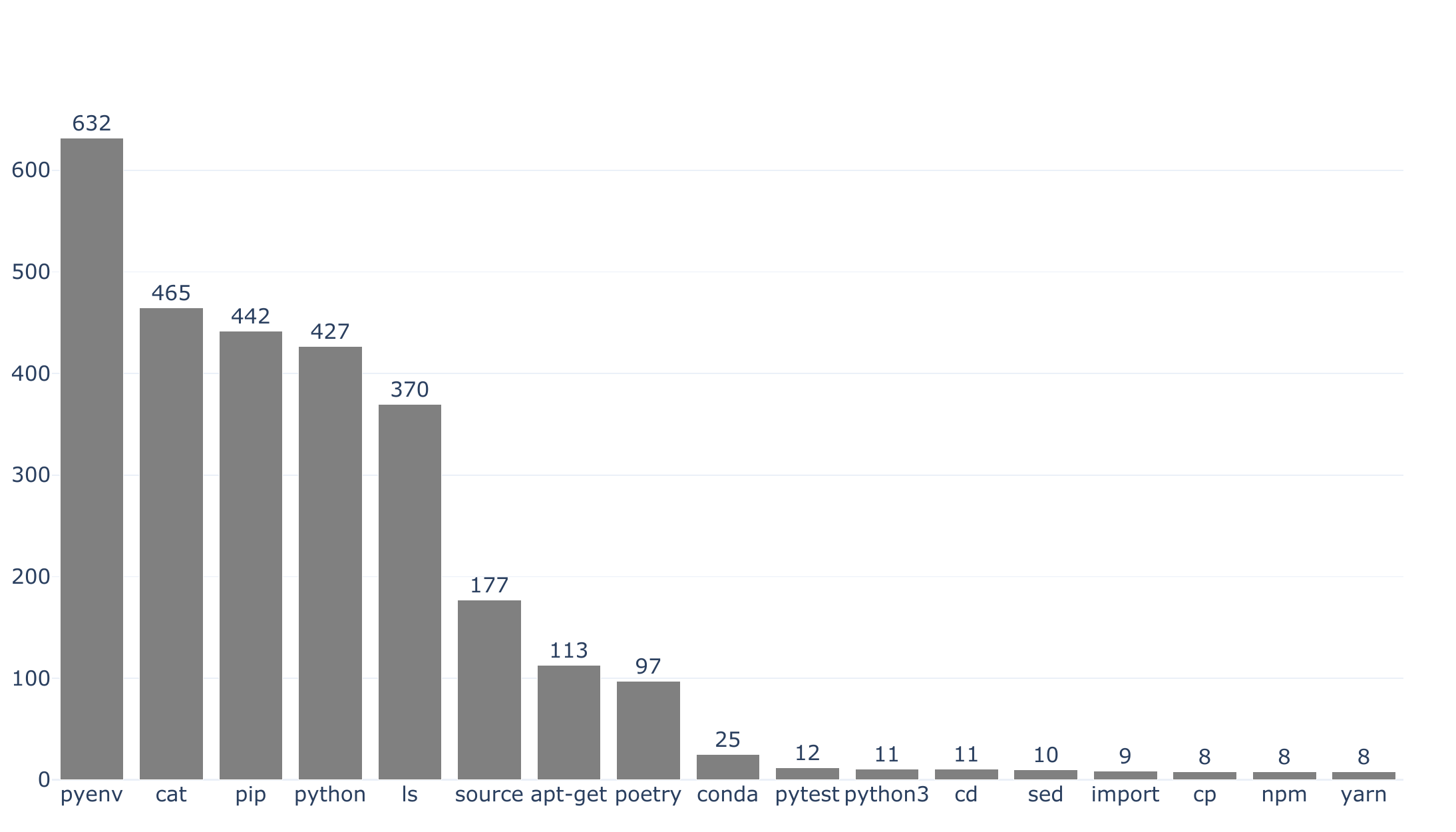}
    \caption{Most frequent Bash commands executed by Bash Agent with GPT-4o on Python dataset.}
    \label{fig:bash_commands_python}
\end{figure}

\begin{figure}[h!]
    \centering
\includegraphics[width=0.8\linewidth]{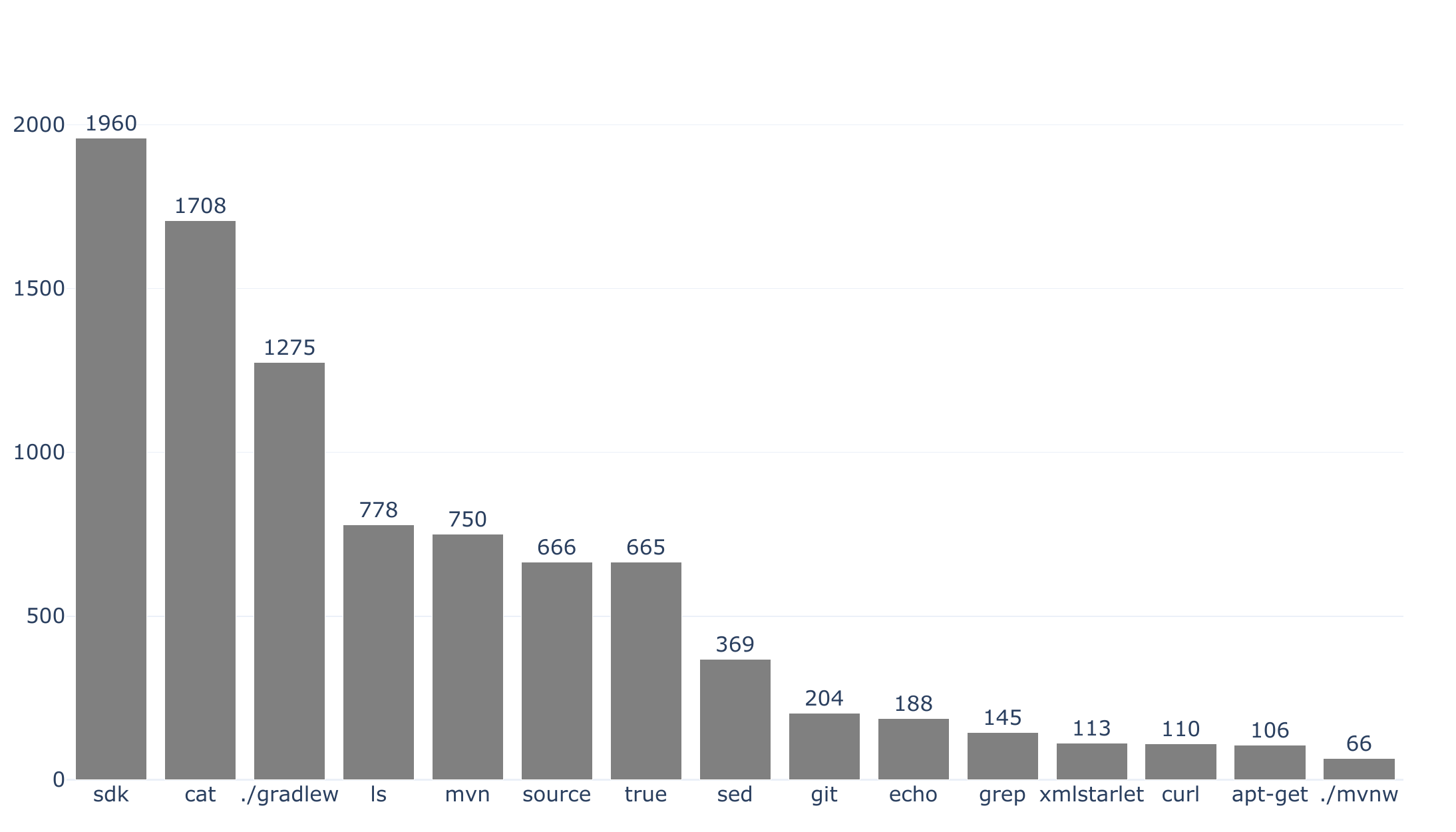}
    \caption{Most frequent Bash commands executed by Bash Agent with GPT-4o on JVM dataset.}
    \label{fig:bash_commands_jvm}
\end{figure}

In~\Cref{tab:script-stats}, we provide additional statistics about the environment setup scripts generated by the considered approaches. The best-performing Bash Agent tends to produce the shortest scripts among considered baselines.
Additionally, for Bash Agent with GPT-4o, we provide the list of the most frequently executed Bash commands across all repositories in our dataset (Python in~\Cref{fig:bash_commands_python}, JVM in~\Cref{fig:bash_commands_jvm}). For both languages, agents actively use file system exploration commands (e.g., \texttt{cat} and \texttt{ls}). There are also a lot of language-specific commands: Python agent uses \texttt{pyenv}, \texttt{pip},  \texttt{python} and \texttt{poetry} a lot, while JVM agent runs \texttt{sdk}, \texttt{./gradlew} and \texttt{maven}.

\section{Expert-produced Scripts}\label{sec:manual}

\begin{table}[h]
    \centering
    \begin{tabular}{llccc}
        \toprule
        \textbf{Baseline} & \textbf{Model} & \textbf{pass@1} &
        \textbf{avgErrs} &
        \makecell{\textbf{Zero}\\\textbf{Exit Code}}\\
        \midrule
        Expert & --- & \textbf{66.7\%}  & \textbf{9.8}& \makecell{\textbf{100\%}\\{\scriptsize 30/30}}\\
        \midrule
        \multirow{2}{*}{Zero-shot LLM}
        &
        GPT-4o  & 10.0\%  & 12.0& \makecell{43\%\\{\scriptsize 13/30}}\\
        & GPT-4o-mini & 6.7\% & 26.1& \makecell{43\%\\{\scriptsize 13/30}}\\
        \midrule
        \multirow{2}{*}{Installamatic Agent} 
        & GPT-4o      & 6.7\%   & 34.9& \makecell{57\%\\{\scriptsize 17/30}}\\
        & GPT-4o-mini & 6.7\%   & 52.9& \makecell{33\%\\{\scriptsize 10/30}}\\
        \midrule
        \multirow{2}{*}{Bash Agent} 
        & GPT-4o              & 0.0\%   & 42.8& \makecell{93\%\\{\scriptsize 28/30}}\\
        & GPT-4o-mini         & 10.0\%  & 16.9& \makecell{93\%\\{\scriptsize 28/30}}\\
        \midrule
        Deterministic Script & --- & 0\%  & 45.9& \makecell{67\%\\{\scriptsize 20/30}}\\
        \bottomrule
    \end{tabular}
    \caption{The results of the environment setup baselines, deterministic script and expert-produced environment setup scripts for 30 randomly sampled Python repositories.}
    \label{tab:manual-results}
\end{table}

\begin{figure}[h]
    \centering
\includegraphics[width=0.9\linewidth]{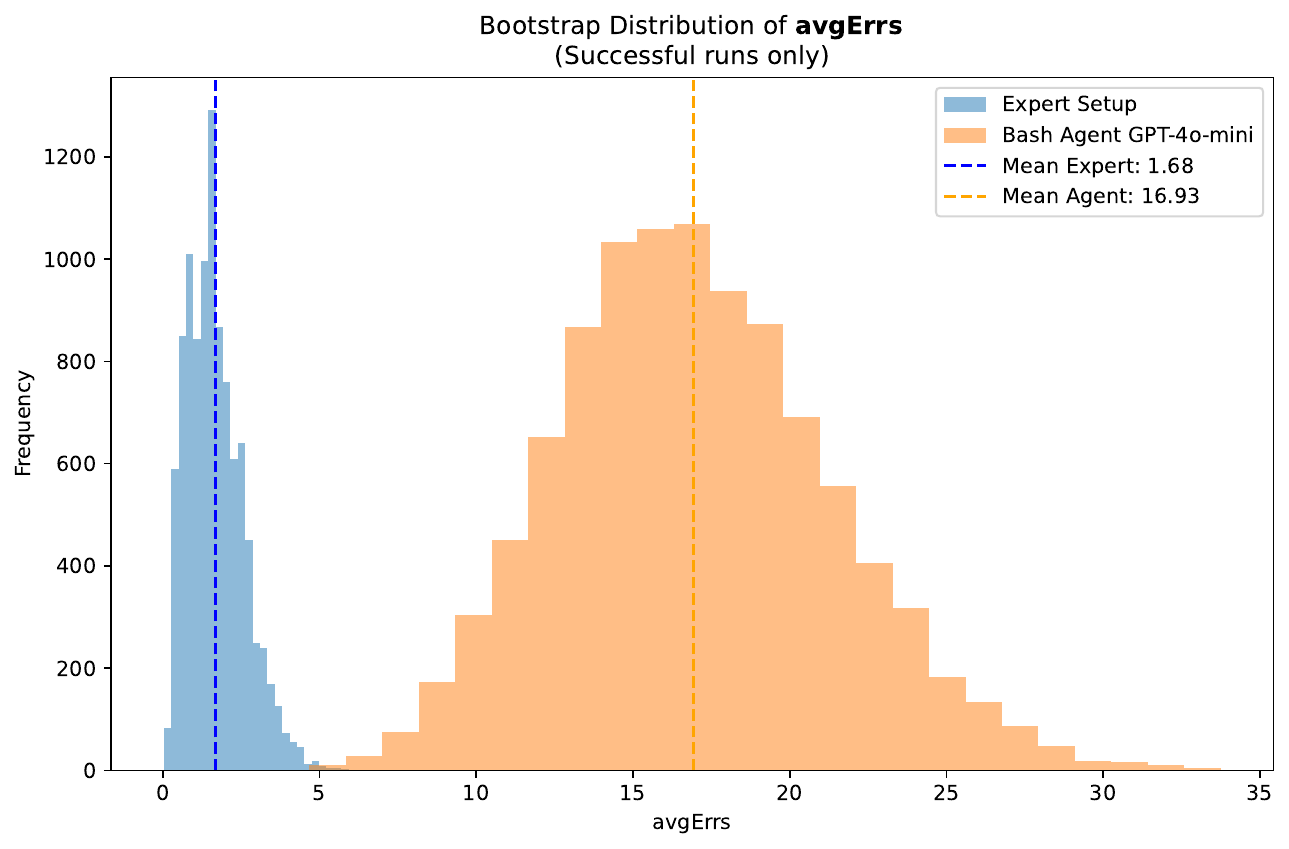}
    \caption{Histograms for \textbf{avgErrs}---the average number of missing errors per repository---for expert-produced scripts and for scripts from Bash Agent with GPT-4o-mini obtained via bootstrap resampling (10,000 iterations).}
    \label{fig:bootstrap-dist}
\end{figure}

We manually investigate the robustness of our proposed environment setup metric for Python on a small sample. 
Specifically, two authors with professional Python software development experience produce curated scripts for 30 randomly selected repositories from our Python sample, and we run our evaluation suite (\Cref{sec:eval-suite}) with those scripts.
The results are presented in~\Cref{tab:manual-results}, and the exact outcomes for each repository are available in~\Cref{tab:manual-results-detailed}. 
For all the considered repositories the expert scripts finished with zero exit code and achieved \textbf{pass@1} of 66.7\%\footnote{Issues that prevented successful setup include the presence of obsolete dependencies (\textit{e.g.}, a legacy Python 2.x module in a Python 3.x codebase) and dynamically resolved imports can't be correctly processed via a static type checker.} and \textbf{avgErrs} of 9.8, outperforming all considered environment setup baselines. Finally, we employ bootstrap resampling (10,000 iterations) to mitigate the small size of the manually processed sample and share the histogram of the distribution of the \textbf{avgErrs} for expert-produced scripts and for Bash Agent with GPT-4o-mini in~\Cref{fig:bootstrap-dist}, which further confirms a wide gap between manually curated scripts and scripts generated by automatic environment setup methods.

\begin{table}
\centering
\resizebox{0.99\textwidth}{!}{
\begin{tabular}{lccccccc}
\toprule
\multirow{2}{*}{\textbf{Repository}} & \multirow{2}{*}{\textbf{Expert}} & \multicolumn{2}{c}{\textbf{Zero-shot LLM}} & \multicolumn{2}{c}{\textbf{Installamatic Agent}} & \multicolumn{2}{c}{\textbf{Bash Agent}}\\
\cmidrule(lr){3-4}\cmidrule(lr){5-6}\cmidrule(lr){7-8}
& & \textbf{GPT-4o-mini} & \textbf{GPT-4o} & \textbf{GPT-4o-mini} & \textbf{GPT-4o} & \textbf{GPT-4o-mini} & \textbf{GPT-4o}\\
\midrule
biopsykit & 2 & - & - & - & - & 4 & 2 \\
cookiecutter & 0 & - & - & - & 60 & 0 & 51 \\
client & 8 & 36 & - & - & - & 36 & 82 \\
mov-cli & 0 & - & 5 & 5 & 5 & 5 & 5 \\
section-properties & 0 & 95 & - & 4 & - & 6 & 6 \\
skrub & 1 & - & 36 & - & 9 & 56 & 96 \\
python-holidays & 0 & 0 & 0 & 0 & 0 & 0 & 14 \\
guardrails & 4 & - & - & - & 41 & 41 & 41 \\
hydra & 25 & 84 & - & 84 & 26 & 25 & 81 \\
duckdb\_engine & 1 & - & - & - & 3 & 3 & 3 \\
pytest-xdist & 0 & - & - & - & 0 & 1 & 1 \\
lobsterpy & 0 & 0 & 0 & - & - & 7 & 8 \\
lhotse & 0 & 61 & - & - & - & 61 & 61 \\
mpmath & 0 & 1 & 1 & - & - & 1 & 1 \\
spectrum-access-system & 2 & - & - & - & - & 39 & 39 \\
adaptix & 0 & - & 0 & 0 & - & 94 & 144 \\
ansible-zuul-jobs & 0 & - & 4 & - & 6 & 6 & 6 \\
photutils & 0 & 5 & - & 142 & - & 2 & 138 \\
stopstalk-deployment & 248 & - & - & - & - & - & - \\
cheroot & 0 & - & - & 27 & - & - & - \\
unixmd & 0 & - & - & - & 50 & 13 & 2 \\
extension-helpers & 0 & - & 3 & - & 3 & 3 & 3 \\
elife-bot & 0 & - & - & - & 265 & 1 & 265 \\
spotpy & 2 & 11 & - & - & - & 8 & 60 \\
bread & 0 & - & 8 & 9 & 9 & 5 & 5 \\
hazm & 2 & - & 79 & - & - & 19 & 19 \\
django-registration & 0 & 1 & 1 & - & 1 & 1 & 1 \\
custodian & 0 & 7 & 9 & 17 & 106 & 9 & 13 \\
lifelines & 0 & 23 & - & 241 & 3 & 0 & 23 \\
smart\_open & 0 & 15 & 10 & - & 7 & 28 & 28 \\
\bottomrule
\end{tabular}}
\caption{The number of missing imports for 30 randomly samples Python repositories for the expert-produced scripts and environment setup baselines.}\label{tab:manual-results-detailed}
\end{table}

\end{document}